# Minimal Neuron Circuits —Part I: Resonators

Amr Nabil, *Member, IEEE*, T. Nandha Kumar, and Haider Abbas F. Almurib, *Senior Member, IEEE*

*Abstract*— Spiking Neural Networks have earned increased recognition in recent years owing to their biological plausibility and event-driven computation. Spiking neurons are the fundamental building components of Spiking Neural Networks. Those neurons act as computational units that determine the decision to fire an action potential. This work presents a methodology to implement biologically plausible yet scalable spiking neurons in hardware. We show that it is more efficient to design neurons that mimic the $I_{Na,p}+I_K$ model rather than the more complicated Hodgkin-Huxley model. We demonstrate our methodology by presenting eleven novel minimal spiking neuron circuits in Parts I and II of the paper. We categorize the neuron circuits presented into two types: Resonators and Integrators. We discuss the methodology employed in designing neurons of the resonator type in Part I, while we discuss neurons of the integrator type in Part II. In part I, we postulate that Sodium channels exhibit type-N negative differential resistance. Consequently, we present three novel minimal neuron circuits that use type-N negative differential resistance circuits or devices as the Sodium channel. Nevertheless, the aim of the paper is not to present a set of minimal neuron circuits but rather the methodology utilized to construct those circuits.

*Impact Statement* — The human brain is estimated to comprise approximately 86 billion neurons, which process information in a highly parallel and efficient nature. To design Spiking Neural Networks that replicate such parallelism and efficiency, a simplistic approach towards neuron circuit design becomes a requirement. Therefore, in the two companion papers, we propose a simplistic biologically plausible approach towards designing minimal neuron circuits. The novel circuits proposed in this work use fewer components than state-of-the-art neuron circuits whilst demonstrating high biological plausibility. This work is intended to serve as a guide for circuit designers, offering a methodology that can be applied to meet various design requirements while ensuring biological plausibility. Additionally, this work aims to enhance the understanding of the analogy between biological neural systems and their electronic counterparts, thereby contributing to the development of more efficient brain-inspired systems.

## I. INTRODUCTION

Spiking Neural Networks (SNNs) have attracted significant attention in recent years [1], primarily due to their event-driven and highly parallel nature. These characteristics make SNNs a potentially more efficient alternative to traditional Artificial Neural Networks (ANNs). SNNs are an emerging subcategory of ANNs that offer a more faithful representation of their biological counterparts. In SNNs, information is conveyed through spikes, also known as action potentials. The information can be represented in the timing or frequency of those spikes [2]. The main building blocks of SNNs are referred to as spiking neurons, which are artificial neurons that aim to faithfully emulate a biological neuron's function. Those neurons accumulate incoming spikes and make the decision to fire an action potential based on the magnitude as well as the frequency of those spikes.

In order to simulate SNNs, several spiking neuron models have been proposed in the literature. Among these models, the Hodgkin-Huxley (HH) [3] model is regarded as the most accurate to date. However, the HH model is a set of four coupled ordinary differential equations, which makes it an inefficient choice for computer simulations. A commonly used alternative is the Leaky Integrate and Fire (LIF) neuron model [4], owing to its simplicity and computational efficiency. However, the LIF model is not biologically plausible since it cannot achieve signal gain, a refractory period or "afterhyperpolarization". These characteristics can be essential for an SNN to deliver satisfactory performance. Rather than simulating an SNN, a more efficient strategy is to design hardware that intrinsically implements it [5]-[9]. In order to exploit this strategy, it is necessary to design neuron circuits that faithfully replicate the qualitative dynamics of those neuron models.

One approach to designing neuron circuits is using transistor-based circuits to implement the models' equations [10]. This approach has been effective in achieving neuron circuits with high biological plausibility. Nevertheless, those circuits use a large number of transistors and capacitors, making them not optimum in terms of scalability. Another approach is to use volatile memristive devices to construct the neuron circuits [11], [12]. The volatile memristor can be used to fulfil the role of a neuronal ion channel in an HH neuron or a threshold-sensing device in an LIF neuron. The volatile memristor can be used in parallel with a capacitor to achieve an LIF neuron [13]-[15]. In this setup, the capacitor resembles the membrane capacitance of the neuron, and the volatile memristor is used to reset the membrane potential after it reaches a certain threshold. In [16], [17], the authors propose neuristor circuits that are able to produce all-or-nothing spiking with signal gain and a refractory period. The authors hypothesize that the Mott memristors used in these circuits can fulfil the role of potassium and sodium ion channels, and therefore, the neuristor circuits might share qualitative similarities with the HH neuron. However, the validity of this hypothesis is disputed in [18].

This work presents a method that can be used to design minimal neuron circuits, whether using transistors, memristive devices, or a hybrid approach that utilizes both. Part I of the paper discusses minimal neuron circuits that act as resonators, while part II discusses ones that act as integrators [19]. We start by introducing the $I_{Na,p}+I_K$ neuron model in section II. Then, in section III, we discuss the nullclines of the $I_{Na,p}+I_K$ neuron model with low threshold Potassium activation, as well as the type of bifurcation the model undergoes to generate spiking. In Section IV, we discuss circuits and devices having type-N Negative Differential Resistance (NNDR), and we show that such circuits and devices can be used to emulate the function of a Sodium channel. Subsequently, we propose three novel minimal neuron circuits of the resonator type in sections V and VI. We use these circuits to show that an NNDR circuit/device can be used along with a MOSFET and RC circuit to realize minimal neuron circuits that implement the $I_{Na,p}+I_K$ model. Additionally, section VII provides a brief discussion comparing our work to other neuron implementations from the literature. Finally, section VIII provides some concluding remarks.

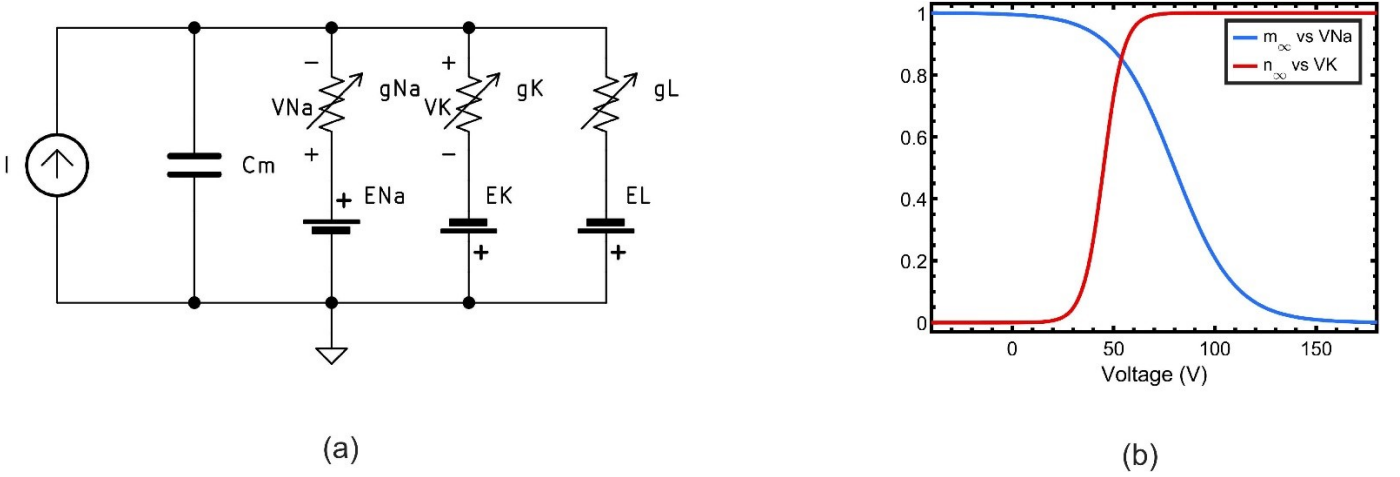

Fig. 1. (a) The $I_{Na,p}+I_K$ model circuit. (b) Steady-state values of the $I_{Na,p}+I_K$ model gating variables as a function of membrane potential.

## II. MINIMAL NEURON MODELS

The HH model is widely accepted as an accurate representation of neuronal dynamics. The model incorporates a neuronal membrane capacitor in parallel with three ion channels. These ion channels play a vital role in enabling the movement of ions across the neuron's membrane. Two channels are voltage-gated, meaning their conductance varies based on the membrane's voltage. Additionally, these two channels exhibit ion selectivity; one manages the flux of sodium ions, while the other manages that of potassium ions. The third channel, referred to as the leak channel, remains unaffected by voltage changes and maintains a constant level of conductance.

Despite the accuracy of the HH model, it is not considered a minimal model for spiking. This means that simpler models can be derived from the HH model while achieving similar qualitative characteristics. A primary consideration in circuit design is to optimize for a minimal number of components. In order to achieve that, we consider the minimal neuron models that can achieve spiking rather than the more complicated HH model. In [20], Izhikevich discusses a set of minimal spiking neuron models, of which the $I_{Na,p}+I_K$ model is of particular interest to this work. The model can be described using a set of only two coupled ordinary differential equations.

$$C\frac{dV}{dt} = I - g_L(V - E_L) - g_{Na}m_\infty(V)(V - E_{Na}) - g_K n(V - E_k) \quad (1)$$

$$\frac{dn}{dt} = \frac{n_\infty(V) - n}{\tau(V)} \quad (2)$$

where $I$ is the injected current, and $C$ is the membrane capacitance. $E_K$, $E_{Na}$, and $E_L$ are voltage sources representing the Nernst potentials of the potassium, sodium and leak ion channels, respectively. The Nernst potential is the potential at which the current flow is zero for a particular ion species. $g_K$, $g_{Na}$ and $g_L$ are the maximum conductance of the potassium, sodium and leak channels, respectively. $n$ is a normalized gating variable for the potassium channel. This gating variable controls the flow of the potassium current. $\tau(V)$ is a time constant that depends on the membrane's voltage. $m_\infty(V)$ and $n_\infty(V)$ are voltage-dependent functions representing the steady state value of the sodium and potassium gating variables, as shown in Fig. 1 (b). Those two functions can be approximated using a Boltzmann function given by

$$n_\infty(V) = \frac{1}{1 + e^{(V_{1/2} - V)/k}} \quad (3)$$

where $V_{1/2}$ is the half-activation voltage satisfying $n_\infty(V_{1/2}) = 0.5$ and $k$ is the slope factor. Two main differences exist between the $I_{Na,p}+I_K$, and HH models. First, the $h$ gating variable is omitted, which can be achieved by considering the relationship $n(t) + h(t) \approx 0.84$ [21]. The second difference is that the sodium gating dynamics are considered to be instantaneous; this assumption can be made since the gating time constant $\tau_m$ of the potassium channel is much larger than the time constant $\tau_n$ of the sodium channel. Therefore, the $I_{Na,p}+I_K$ model becomes a two-dimensional system.

The $I_{Na,p}+I_K$ model is not qualitatively or quantitatively equivalent to the HH model. Nevertheless, this model's mechanism of action potential generation closely resembles that of the HH model. The mechanism of action potential generation can be explained by analyzing the interplay of the neuronal gating variables. If a small current pulse is injected into the $I_{Na,p}+I_K$ model, the membrane capacitor would start charging, and the membrane potential would gradually increase.

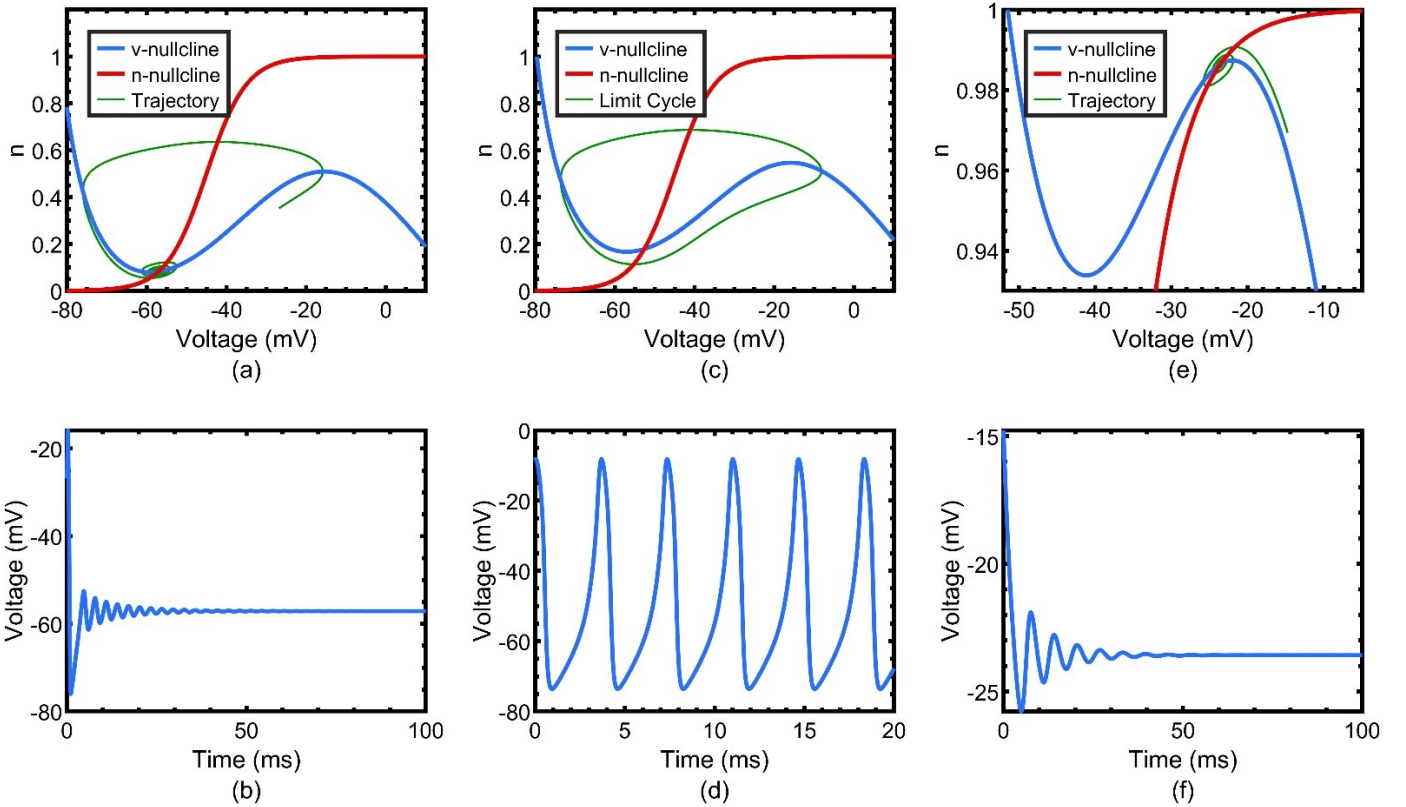

Fig. 2. Nullcline analysis (a,c,e) and voltage output (b,d,f) of the $I_{Na,p}+I_K$ model with low Potassium threshold with $C = 1\ mF$, $E_L = -78\ mV$, $E_{Na} = 60\ mV$, $E_K = -90\ mV$, $g_L = 8\ S$, $g_{Na} = 20\ S$, $g_K = 10\ S$, $V_{1/2_{Na}} = -20\ mV$, $V_{1/2_K} = -45\ mV$, $k_{Na} = 15*10^{-3}$, $k_K = 5*10^{-3}$ and $\tau = 1\ ms$. (a,b) I = 12 mA. (c,d) I = 40 mA. (e,f) I = 354 mA [20].

However, the potassium gate would open soon after, and $E_K$ would pull the membrane potential back towards the resting value. Therefore, no action potentials would be generated.

On the contrary, if a sufficiently large current pulse is injected, the membrane potential would increase significantly. Since the sodium channel activation is instantaneous, the sodium channel would start opening, further amplifying the increase in the membrane potential. This positive feedback effect results in the upstroke of the membrane potential. On the other hand, the potassium channel would require time to start opening due to $\tau_n$. Once the potassium channel opens, the membrane potential is pulled back towards the resting potential, resulting in the downstroke of the action potential. Therefore, the sodium channel would close, but the potassium channel would remain open for a period dictated by $\tau_n$. This would result in the membrane potential falling below the resting potential, a phenomenon referred to as "afterhyperpolarization". When the Potassium channel closes, the injected current would start charging the membrane capacitor again, and the same process would be repeated. Therefore, sustained action potentials would be generated through this mechanism. All the circuits in this work draw inspiration from this model and have a similar action potential generation mechanism.

### III. $I_{NA,P}+I_K$ MODEL ($I_K$ WITH LOW THRESHOLD)

Izhikevich classifies neurons into two types, namely resonators and integrators, depending on their qualitative dynamics [20]. Resonators are neurons that show subthreshold oscillations and do not have a clear-cut threshold for firing action potentials. On the other hand, neurons classified as integrators have a well-defined threshold for firing action potentials and display all-or-nothing spiking behaviour. Another significant difference between the two types is that integrators prefer high-frequency inputs; the higher the frequency of the input stimuli, the sooner the neuron fires an action potential. While resonators fire action potentials in response to inputs with frequencies similar to that of their subthreshold oscillations, frequencies higher than that are less likely to generate an action potential. The type of the neuron can be determined based on the type of bifurcation the neural system undergoes. Neurons that undergo an Andronov-Hopf bifurcation can be classified as resonators, while those that undergo a saddle-node bifurcation can be classified as integrators.

The $I_{Na,p}+I_K$ model discussed in the previous section can have a Potassium channel with either a low or high threshold for activation. Each of those two choices results in significantly different qualitative dynamics since a low threshold results in a neuron that can be classified as a resonator, while a high threshold results in one that can be classified as an integrator. This section discusses the $I_{Na,p}+I_K$ model with low threshold potassium activation. We plot the system's nullclines and discuss the type of bifurcation the system undergoes to generate spiking. To obtain the v-nullcline and n-nullcline of the system, we set $\frac{dV}{dt}$ and $\frac{dn}{dt}$ in equations 1 and 2 to zero. Therefore, the v-nullcline and n-nullcline are given by equations 4 and 5, respectively.

4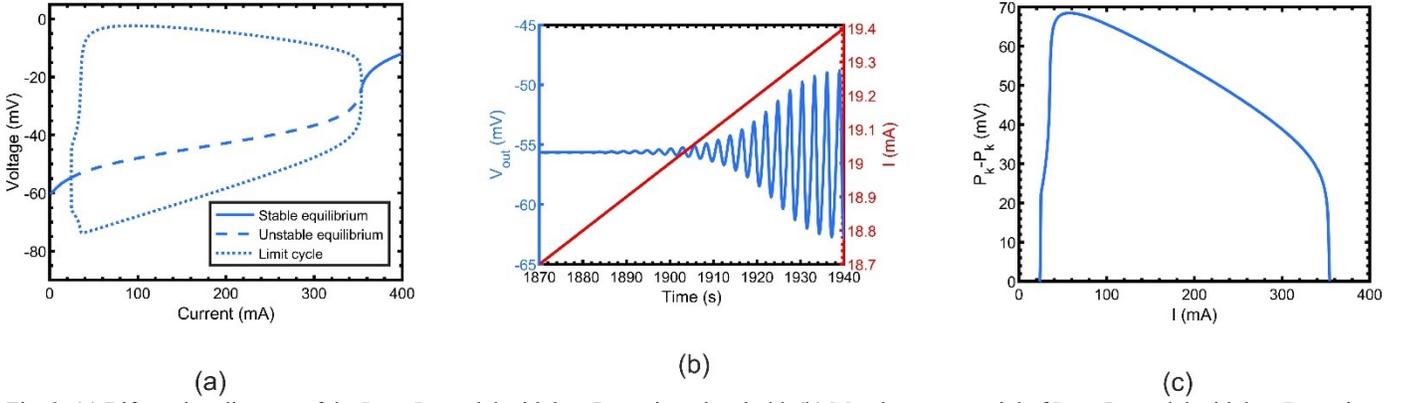

(a)  (b)  (c)

Fig. 3. (a) Bifurcation diagram of the $I_{Na,p}+I_K$ model with low Potassium threshold. (b) Membrane potential of $I_{Na,p}+I_K$ model with low Potassium threshold in response to a ramp input. (c) Peak to peak amplitude of the membrane potential oscillations of the $I_{Na,p}+I_K$ model with low Potassium threshold as a function of $I$. The same parameter values in Fig. 2 were used.

$$n = \frac{I - g_L(V - E_L) - g_{Na}m_\infty(V)(V - E_{Na})}{g_K(V - E_K)} \quad (4)$$

$$n = n_\infty(V) \quad (5)$$

Fig. 2 shows the nullclines of the system with varying injected current $I$ along with the corresponding output waveform of the membrane potential. The v-nullcline in Fig. 2 is an inverted N-shaped curve, while the n-nullcline is a sigmoid curve following $n_\infty(V)$. Fig. 2 (a) shows the nullclines for an injected current of $12\ mA$. The nullclines intersect towards the left branch of the v-nullcline, and the equilibrium is a stable focus. The green trajectory in Fig. 2 (a) converges towards the stable focus and rotates around the equilibrium as it approaches it. These rotations correspond to the subthreshold oscillations in Fig. 2 (b), where the membrane potential converges to a resting value and no spiking is observed.

If the injected current is increased to $40\ mA$, the equilibrium moves towards the middle branch of the v-nullcline. This results in the stable focus losing stability through a supercritical Andronov-Hopf bifurcation. Consequently, this gives birth to a stable limit cycle around the now unstable focus. The limit cycle corresponds to the sustained spiking activity shown in Fig. 2 (d). Interestingly, when the injected current is further increased to $354\ mA$, the neural system undergoes another supercritical Andronov-Hopf bifurcation. The limit cycle shrinks and disappears, and the focal equilibrium regains stability. As illustrated in Fig. 2 (e), the green trajectory converges to the stable focus, and the membrane potential settles at a resting value, as shown in Fig. 2 (f). This occurs because the injected current becomes too large; thus, the current generated by the fully open Potassium channel becomes insufficient to pull the membrane potential down.

Fig. 3 (a) shows the bifurcation diagram of the $I_{Na,p}+I_K$ model with a low Potassium threshold, where the injected current $I$ is used as the bifurcation parameter. The figure shows that at injected currents lower than approximately $20\ mA$, the circuit has one stable equilibrium, and therefore, the membrane potential always converges to a resting value. As the injected current increases, the equilibrium loses stability, and a stable limit cycle is generated through a supercritical Andronov-Hopf bifurcation. This implies that the membrane potential shows sustained oscillations for currents larger than $20\ mA$. Nevertheless, any current greater than $350\ mA$ will not result in spiking since the limit cycle disappears via another supercritical Andronov-Hopf bifurcation, and the equilibrium becomes stable again.

Izhikevich discusses the response of neurons undergoing various types of bifurcation to a ramp injected current [20]. He shows that neurons that exhibit spiking through a supercritical Andronov-Hopf bifurcation show oscillations that commence with near zero amplitude, and the amplitude increases as the injected current increases. Fig. 3 (b) shows the response of the $I_{Na,p}+I_K$ model with a low threshold to a ramp injected current near the bifurcation value. The oscillations in fig. 3 (b) start with minimal amplitude, and the amplitude increases as the injected current increases, which is consistent with Izhikevich's postulation. This is one of the distinctive features of neurons undergoing a supercritical Andronov-Hopf bifurcation.

Fig. 3 (c) shows the peak to peak amplitude of the membrane potential oscillations as a function of the bifurcation parameter $I$. The oscillations commence at $I > 20\ mA$ with near zero amplitude, and increases sharply as the injected current is increased. This sharp increase in amplitude can also be inferred from the bifurcation diagram in Fig. 3 (a), where the magnitude of the limit cycle increases rapidly as the current is increased. Nevertheless, the peak to peak amplitude starts to decrease as the injected current is increased further, which is attributed to the supercritical Andronov-Hopf bifurcation that occurs at $I \approx 354\ mA$. At $I \approx 354\ mA$, the amplitude of spiking approaches a zero value, and the spiking disappears if the current is increased beyond that value.

## IV. TYPE-N NEGATIVE DIFFERENTIAL RESISTANCE DEVICES

NNDR devices/circuits are devices/circuits that exhibit voltage-controlled negative differential resistance in their I-V characteristics, i.e. the slope is negative for a portion of the I-V characteristics. Additionally, the I-V characteristics is an N-shaped curve, hence the name "Type-N". From the characteristics in Fig. 1 (b), we can infer that the sodium ion channel actually possesses this NNDR property. The NNDR property is the core qualitative characteristic that contributes to spiking. This is because the NNDR property allows the sodium channel to amplify voltage changes through a positive feedback effect, which is necessary to generate the upstroke of the spike. Therefore, to find a device/circuit that can emulate the function of a sodium channel, that device/circuit must show NNDR. Even if the device/circuit's characteristics differ from that of the sodium channel, the device/circuit could still potentially mimic a Sodium channel if it exhibits the NNDR property.

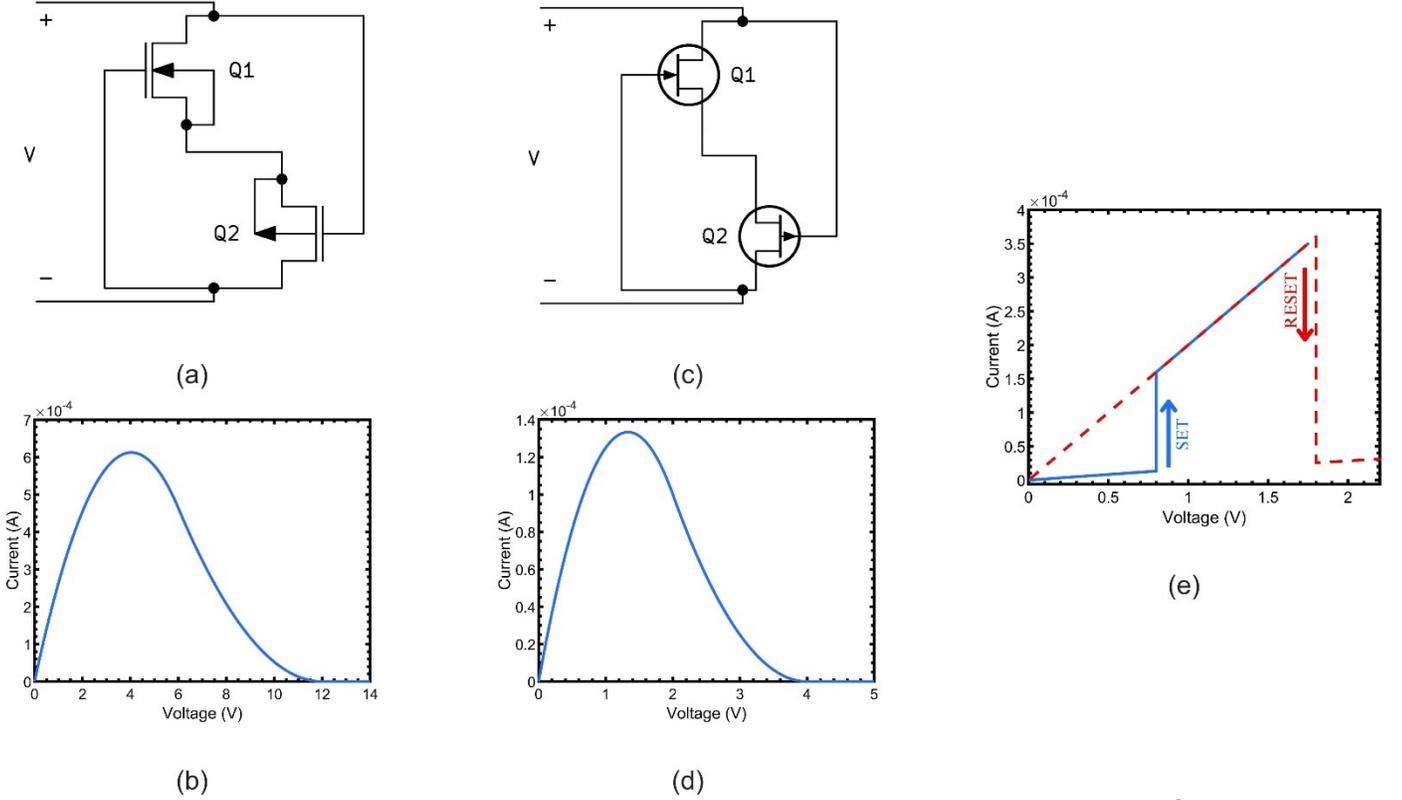

Fig. 4. (a), (b) NNDR circuit comprised of two complementary MOSFETs and its I-V characteristics. $K_{n1} = 100\ uA/V^2$, $V_{t01} = -6\ V$, $\lambda_1 = 0.01$, $K_{p2} = 100\ uA/V^2$, $V_{t02} = 6\ V$ and $\lambda_2 = 0.01$. (c), (d) NNDR circuit comprised of two complementary JFETs and its I-V characteristics. $\beta_1 = 100\ uA/V^2$, $V_{t01} = -2\ V$, $\lambda_1 = 0$, $\beta_2 = 100\ uA/V^2$, $V_{t02} = -2\ V$ and $\lambda_2 = 0$. (e) I-V characteristics of the unipolar memristor model, with $R_{on} = 5\ K\Omega$, $R_{off} = 70\ K\Omega$, $\alpha = 5*10^{10}$, $\beta = 10^{10}$, $V_{rst} = 1.8\ V$ and $V_{set} = 0.8\ V$.

In [22], Leon O. Chua presents multiple NNDR circuits that do not require an internal power supply. These circuits are comprised of a minimal number of either BJT, JFET or MOSFET transistors or a combination of them. Fig. 4 (a) and Fig. 4 (c) show two of those NNDR circuits using two MOSFETs and two JFETs, respectively. The I-V characteristics of the circuits shown in Fig. 4 (b) and Fig. 4 (d) show voltage-controlled negative differential resistance, i.e. the current decreases as voltage increases for a portion of their characteristics. Interestingly, some unipolar resistive switching devices have a RESET voltage larger than the SET voltage [23], [24], [25], [26] and, therefore, can be considered NNDR devices. One of the main aims of this work is to show that NNDR devices and circuits can be used as sodium channels for neuron circuits. In order to simulate unipolar resistive switching devices, we use a simple model [27] described using equations 6 and 7.

$$I_m = r^{-1} V_m \quad (6)$$

$$\frac{dr}{dt} = \begin{cases} \alpha |V_m| H(R_{off} - r), & |V_m| > V_{rst} \\ \beta |V_m| H(r - R_{on}), & V_{set} \leq |V_m| < V_{rst} \\ 0, & Otherwise \end{cases} \quad (7)$$

where $I_m$ and $V_m$ are the current through the device and voltage across the device, respectively. $r$ is the device's resistance and the model's state variable. $V_{rst}$ and $V_{set}$ are the voltage thresholds for the RESET and SET processes, respectively. $\alpha$ and $\beta$ are fitting constants that control the speed of the RESET and SET processes, respectively. $H(x)$ is the Heaviside step function. $R_{off}$ and $R_{on}$ are the maximum and minimum resistance of the device, respectively.

We modified the model in [27] considering that $V_{rst}$ is larger than $V_{set}$ since, in this work, we only consider unipolar resistive switching devices that exhibit NNDR. Therefore, if a voltage larger than $V_{set}$ and smaller than $V_{rst}$ is applied to the device, the resistance of the device decreases with a rate determined by $\beta |V_m|$. Similarly, if a voltage larger than $V_{rst}$ is applied to the device, the resistance increases with a rate determined by $\alpha |V_m|$. The Heaviside step functions are used to limit the resistance between $R_{off}$ and $R_{on}$. The I-V characteristics of the unipolar memristor model is shown in Fig. 4 (e). In section VI, we use this unipolar memristor model to show that a unipolar memristor with $V_{rst}$ larger than $V_{set}$ can effectively be used to mimic a Sodium channel.

V. FET-BASED MINIMAL NEURON (RESONATOR)

We propose a novel minimal neuron circuit that comprises three MOSFETs, two capacitors and a resistor, as shown in Fig. 5. The circuit uses significantly fewer components than other neuron implementations [28], [29], yet it exhibits spiking with a mechanism similar to that of the $I_{Na,p}+I_K$ model with low threshold Potassium activation. The circuit comprises a Potassium and Sodium channel along with the membrane capacitor denoted as $C_1$. The Potassium channel is comprised of $R_1$, $C_2$ and $Q_1$. $R_1$ and $C_2$ form an RC circuit that determines the time constant for the activation of the Potassium channel. For the circuit to be of the resonator type, the threshold $V_{t01}$ of $Q_1$ has to be low so that the MOSFET's channel would be conducting at rest. This is to mimic the Potassium channel of the $I_{Na,p}$+$I_K$ model discussed in section III, where the channel has a low threshold and is partially open at rest. On the other hand, the NNDR circuit comprising $Q_3$ and $Q_4$ acts as the sodium channel, which is considered to have instantaneous activation as in the $I_{Na,p}$+$I_K$ model. In order for the neuron circuit to be excitable, the NNDR circuit must be biased in the negative differential region of its I-V characteristics. Using the parameters in Fig. 6, the resting value of $V_{out}$ is $234\ mV$ for zero injected current $I$. With $V_{dc}$ equal to $3.5\ V$, the voltage across the NNDR circuit is $3.266\ V$. This ensures that the NNDR circuit operates in the negative differential resistance region shown in Fig. 4 (b).

In Fig. 6, we plot the circuit's nullclines along with the corresponding membrane potential output with varying injected current $I$. The nullclines are obtained by first finding the circuit's equations given by equations 8 and 9 and equating the derivatives of the state variables to zero.

$$\frac{dV_{out}}{dt} = \frac{1}{C_1}\left(I - \frac{V_{out} - V_{GS}}{R_1} - i_{DSQ_1} + i_{DSQ_3}\right) \quad (8)$$

$$\frac{dV_{GS}}{dt} = \frac{1}{C_2 R_1}(V_{out} - V_{GS}) \quad (9)$$

The state variables used are $V_{out}$ and $V_{GS}$, which denote the voltage across the capacitors $C_1$ and $C_2$, respectively. $i_{DSQ_1}$ is the current through the channel of the MOSFET $Q_1$ and resembles the potassium current of the $I_{Na,p}$+$I_K$ model. Similarly, $i_{DSQ_3}$ is the current through the channels of the MOSFETs $Q_2$ and $Q_3$ and mimics the sodium current of the $I_{Na,p}$+$I_K$ model. The Shichman and Hodges equations [30] are used to describe the MOSFETs and obtain the currents $i_{DSQ_1}$ and $i_{DSQ_3}$ numerically since an explicit solution is not possible due to the discontinuous nature of those equations.

Fig. 6 (a) shows the $V_{out}$ and $V_{GS}$ nullclines of the circuit with an injected current of $2\ uA$. The $V_{out}$ nullcline is an inverted N-shape curve similar to the V-nullcline of the $I_{Na,p}$+$I_K$ model. With an injected current of $2\ uA$, the nullcines' intersection is a stable focus equilibrium. This can be inferred from the green trajectory converging towards the focal equilibrium while making diminishing rotations. Those rotations correspond to the subthreshold oscillations observed in Fig. 6 (b) and $V_{out}$ approaching a resting value. The subthreshold oscillations are due to an interplay between the current through $Q_1$ (Potassium current) and the current through $Q_2$ and $Q_3$ (Sodium Current). The injected current charges $C_1$ and $V_{out}$ increases. However, since $Q_1$ has a low threshold and its channel is partially conducting at rest, the current through $Q_1$ increases and pulls down $V_{out}$ before generating a spike. Due to the slow time constant for discharging $C_2$, $V_{out}$ continues to decrease below the resting value. Once $C_2$ discharges sufficiently, $V_{GS}$ decreases, which causes $Q_1$'s channel to become less conducting and its current to decrease. Therefore, the injected current starts charging $C_1$ again. This process results in the subthreshold oscillations shown in Fig. 6 (b) and is qualitatively equivalent to that of the $I_{Na,p}$+$I_K$ model discussed in section II.

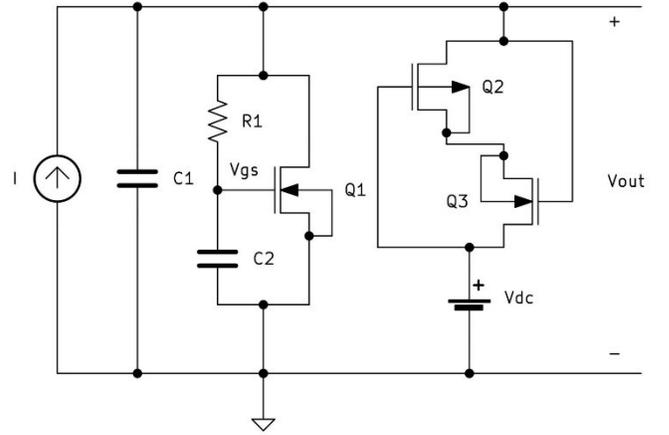

Fig. 5. FET-based minimal neuron circuit. (Resonator)

If the injected current increases to $70\ uA$, the circuit undergoes a supercritical Andronov-Hopf bifurcation. The focal equilibrium loses stability while giving birth to a stable limit cycle. This is evident from the green trajectory shown in Fig. 6 (c), where the trajectory follows a periodic orbit. This limit cycle corresponds to the sustained spiking behaviour shown in Fig. 6 (d). The type of bifurcation here is similar to that of the $I_{Na,p}$+$I_K$ model discussed in section III, which shows the qualitative similarity between the two systems. The spikes in Fig. 6 (d) consist of an upstroke followed by a downstroke with $V_{out}$ going below the resting value. The shape of the spike is qualitatively similar to that of the $I_{Na,p}$+$I_K$ model with a low threshold shown in Fig. 2 (d).

Additionally, the spike generation mechanism is analogous to that of the $I_{Na,p}$+$I_K$ model. The injected current $I$ charges $C_1$, and therefore $V_{out}$ increases. This causes the voltage across the NNDR circuit ($Q_2$ and $Q_3$) to decrease. Consequently, the current through the NNDR circuit increases, which further amplifies the increase in $V_{out}$ and results in the upstroke of the spike. This positive feedback effect is due to the negative differential resistance of the NNDR circuit and is akin to the one observed in the sodium channel of the $I_{Na,p}$+$I_K$ model. During the upstroke, $C_2$ charges and $V_{GS}$ increases. However, due to the slow time constant for charging $C_2$, the current through the channel of $Q_1$ does not increase before the upstroke is generated. Once $C_2$ charges sufficiently, the current through the channel of $Q_1$ increases, which pulls down $V_{out}$ and results in the downstroke of the spike. Since the time constant for discharging $C_2$ is slow, $V_{out}$ continues to decrease below the resting value, resulting in "afterhyperpolarization", similar to the one observed in the $I_{Na,p}$+$I_K$ model. When $C_2$ discharges, the current through $Q_1$ decreases and the injected current $I$ starts charging $C_1$ again, resulting in the sustained spiking shown in Fig. 6 (d).

Interestingly, if we increase the injected current to $150\ uA$, the circuit undergoes another supercritical Andronov-Hopf bifurcation. The limit cycle shrinks and disappears, and the focal equilibrium regains stability. This is evident from the green trajectory shown in Fig. 6 (e) and the equivalent voltage output waveform in Fig. 6 (f). In this case, the channel of $Q_1$ is fully conducting. However, since the injected current is too large, the current through $Q_1$'s channel becomes insufficient to pull down $V_{out}$. Therefore, a resting state is achieved through a balance between the injected current and the currents through the channels of $Q_1$, $Q_2$ and $Q_3$.

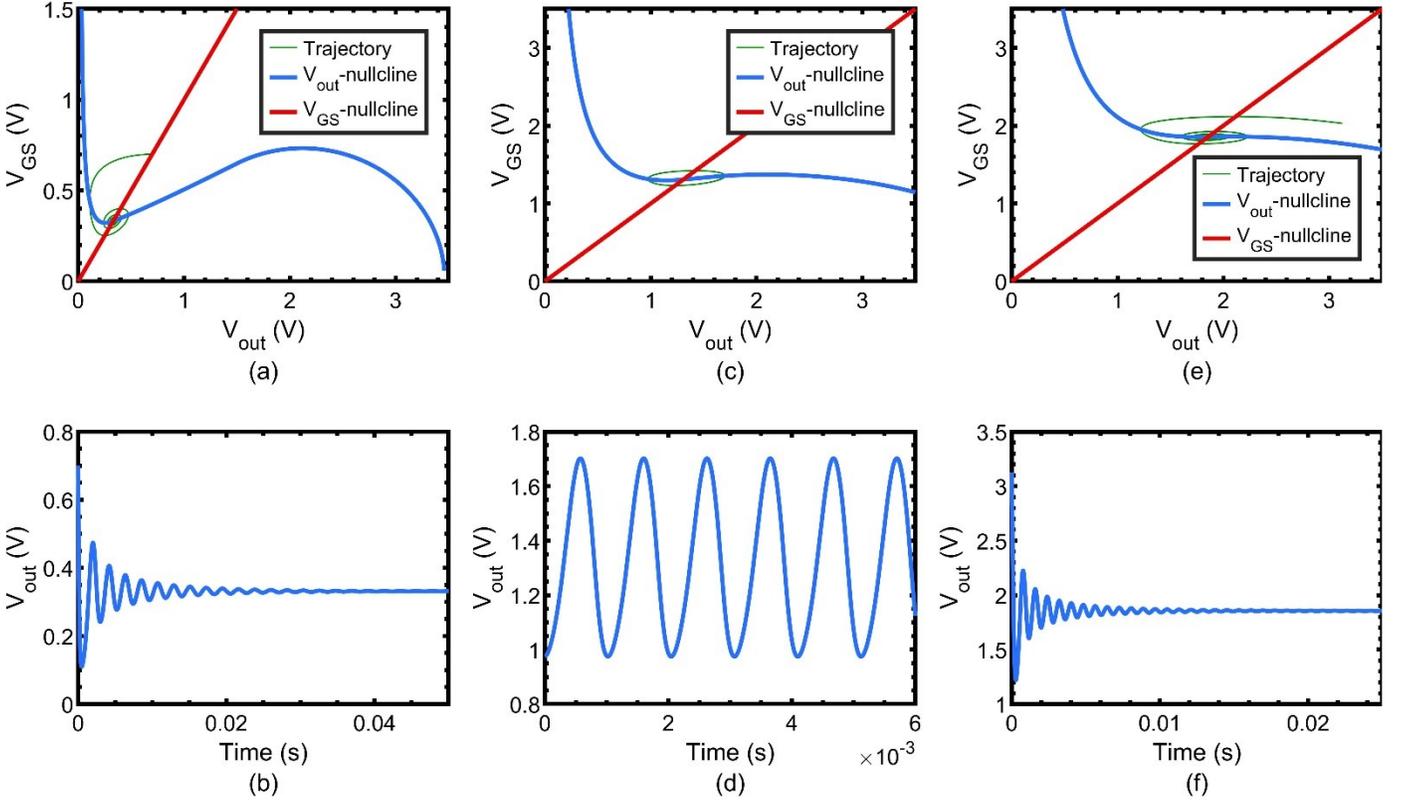

Fig. 6. Nullcline analysis (a,c,e) and voltage output (b,d,f) of the FET-based minimal neuron circuit (Resonator) in Fig. 5 with $C_1 = 5\ nF$, $C_2 = 0.6\ nF$, $R_1 = 1\ M\Omega$, $V_{dc} = 3.5\ V$, $K_{n1} = 100\ uA/V^2$, $V_{t01} = 0\ V$, $\lambda_1 = 0.01$, $K_{p2} = 40\ uA/V^2$, $V_{t02} = 2\ V$, $\lambda_2 = 0.01$, $K_{n3} = 40\ uA/V^2$, $V_{t03} = -2\ V$ and $\lambda_3 = 0.01$. (a,b) I = 2 uA. (c,d) I = 70 uA. (e,f) I = 150 uA.

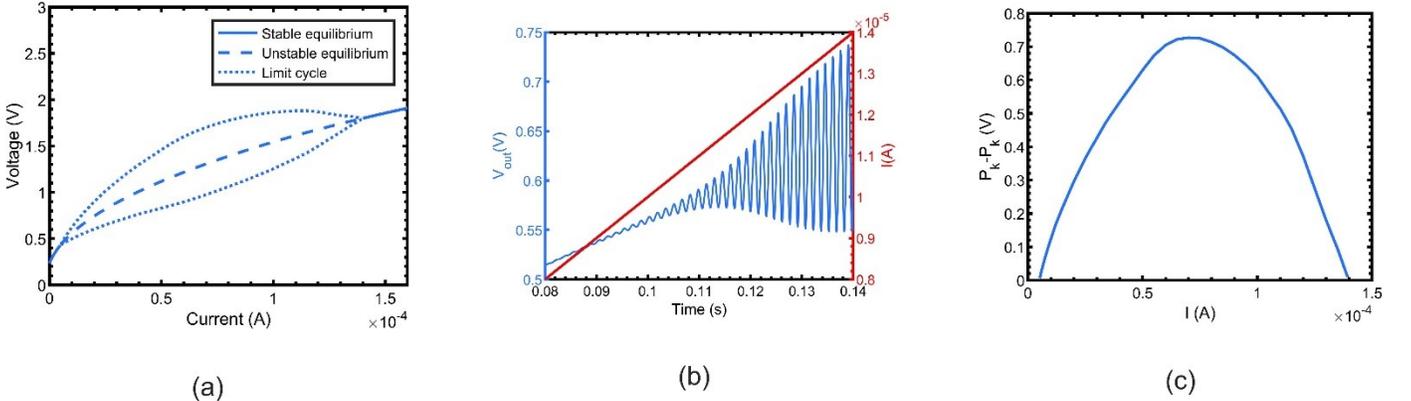

Fig. 7. (a) Bifurcation diagram of the FET-based minimal neuron circuit in Fig. 5. (b) Membrane potential of the FET-based minimal neuron circuit in response to a ramp input. (c) Peak to peak amplitude of the membrane potential oscillations of the circuit in Fig. 5 as a function of $I$. The same parameter values in Fig. 6 were used.

Fig. 7 (a) shows the bifurcation diagram of the circuit with the injected current $I$ as the bifurcation parameter. Currents below $5\ uA$ result in the circuit having one stable focal equilibrium, and therefore, no spiking is produced, and $V_{out}$ converges towards the resting value of the equilibrium. If the current increases beyond $5\ uA$, the circuit undergoes a supercritical Andronov-Hopf bifurcation. This results in the appearance of a stable limit cycle and the focal equilibrium losing stability, as shown in Fig. 7 (a). Therefore, the circuit shows sustained spiking for currents larger than $5\ uA$, as illustrated using the example in Fig. 6 (d). However, the circuit undergoes another supercritical Andronov-Hopf bifurcation at about $140\ uA$, which results in the focal equilibrium regaining stability and the limit cycle disappearing. Therefore, currents beyond $140\ uA$ will not result in spiking, and $V_{out}$ will always approach a resting value. The bifurcation diagram is qualitatively similar to that of the $I_{Na,p}+I_K$ with a low threshold Potassium activation shown in Fig. 3 (a).

Fig. 7 (b) shows the circuit's response to a ramp injected current $I$ near the bifurcation. The output voltage $V_{out}$ shows oscillations with near zero amplitude, and the amplitude starts increasing as the injected current $I$ increases. This is similar to the response observed in Fig. 3 (b) for the $I_{Na,p}+I_K$ model with a low threshold, which is in agreement with Izhikevich's postulation discussed in section III [20].

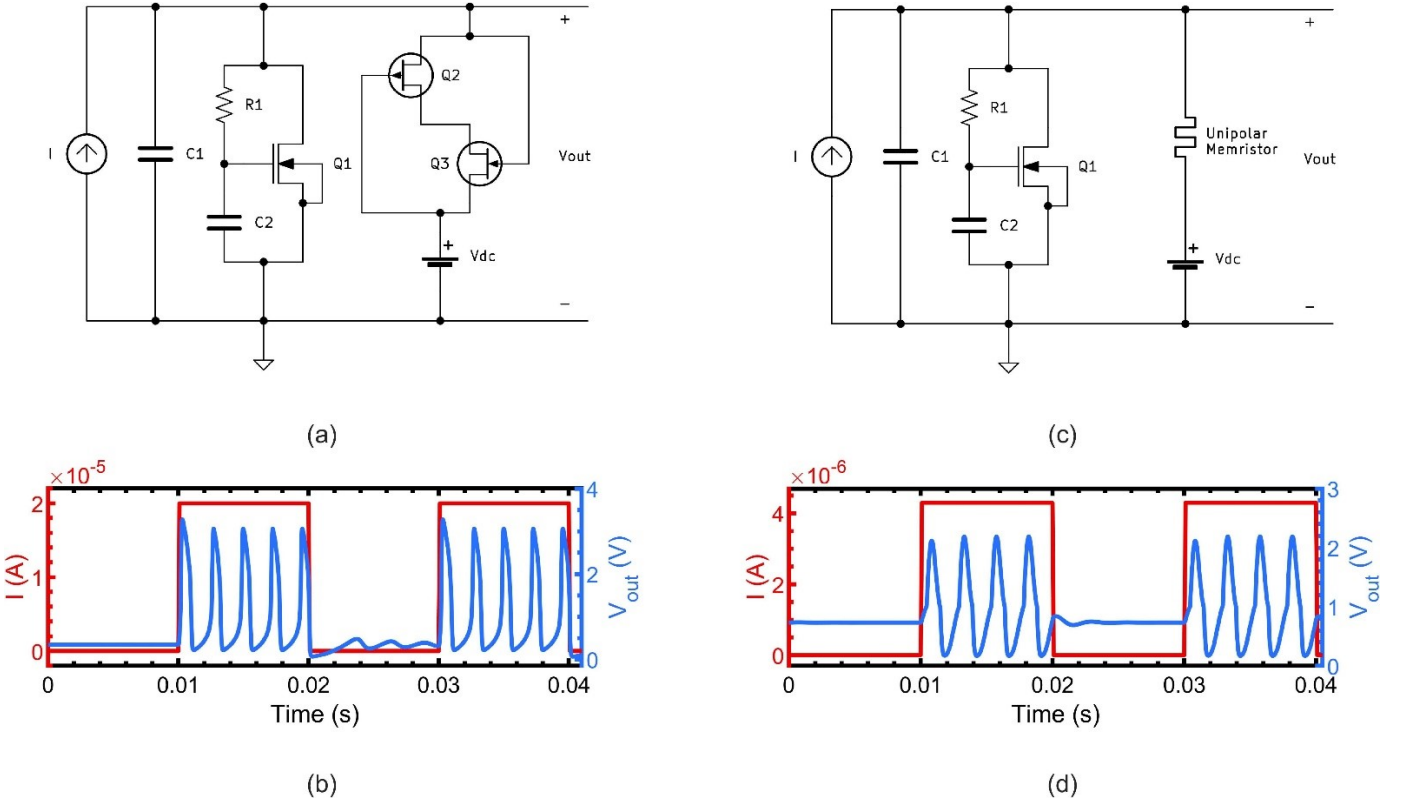

Fig. 8. Two minimal neuron circuits (Resonators) with their corresponding membrane potential output waveform. (a), (b) $C_1 = 5\ nF$, $C_2 = 1\ nF$, $R_1 = 1\ M\Omega$, $V_{dc} = 3.5\ V$, $K_{n1} = 40\ uA/V^2$, $V_{t01} = -0.5\ V$, $\lambda_1 = 0.01$, $\beta_2 = 130\ uA/V^2$, $V_{t02} = -2\ V$, $\lambda_2 = 0$, $\beta_3 = 130\ uA/V^2$, $V_{t03} = -2\ V$ and $\lambda_3 = 0$. (c), (d) $C_1 = 5\ nF$, $C_2 = 1\ nF$, $R_1 = 1\ M\Omega$, $V_{dc} = 4\ V$, $K_{n1} = 120\ uA/V^2$, $V_{t01} = -0.5\ V$, $\lambda_1 = 0.01$, $R_{on} = 100\ K\Omega$, $R_{off} = 1\ M\Omega$, $\alpha = 5*10^{10}$, $\beta = 10^{10}$, $V_{rst} = 3\ V$ and $V_{set} = 1\ V$.

Fig. 7 (c) shows the peak to peak amplitude of the membrane potential oscillations as a function of the injected current $I$. As shown in Fig. 7 (c), the oscillations commence with near zero amplitude at $I \approx 5\ uA$ (bifurcation value), and the amplitude of oscillation increases gradually as $I$ is increased. After the amplitude of oscillation reaches its peak, the amplitude starts to decrease as $I$ is increased further. This decrease in amplitude is due to the circuit approaching the supercritical Andronov-hopf bifurcation that results in the disappearance of the limit cycle. This behavior can also be inferred from the bifurcation diagram in Fig. 7 (a). Fig. 3 (c) and Fig. 7 (c) are quantitatively different, but share qualitative similarities, since the increase and decrease in amplitude of oscillations in both cases are attributed to a supercritical Andronov-Hopf bifurcation that results in the appearance or disappearance of a stable limit cycle.

## VI. OTHER MINIMAL NEURONS (RESONATORS)

We present another two novel minimal neuron circuits in Fig 8 (a) and Fig 8 (c). The difference between these circuits and the circuit in Fig. 5 is that the NNDR circuit in Fig. 5 is replaced with an NNDR circuit comprised of two complementary JFETs in Fig. 8 (a) and a unipolar memristor in Fig. 8 (c). This shows that many NNDR devices/circuits can effectively be used as Sodium channels. Both circuits demonstrate sustained oscillations with signal gain and "afterhyperpolarization", as shown in Fig. 8 (b) and Fig 8 (d). Both circuits can also be classified as resonators since they show subthreshold oscillations. The spike generation mechanism of the circuit in Fig. 8 (a) is identical to that of the circuit discussed in section V and, therefore, will not be discussed further in this section.

In order for the circuit in Fig. 8 (c) to demonstrate excitability, the unipolar memristor must be biased in the NNDR region of its IV characteristics shown in Fig. 4 (e), i.e. the voltage across the memristor should be larger than $V_{rst}$, and the memristor should be at the High Resistance State (HRS). The circuit is able to demonstrate subthreshold oscillations through an interplay between the injected current and the slow current through $Q_1$'s channel. The injected current $I$ charges $C_1$ and $C_2$, which causes $V_{out}$ and $V_{gs}$ to increase. As $V_{gs}$ increases, the current through $Q_1$'s channel increases as well. Thus, $V_{out}$ decreases towards the resting potential. However, due to the slow time constant for discharging $C_2$, $V_{out}$ continues to decrease even below the resting potential. Once $C_2$ discharges the injected current starts charging $C_1$ again, resulting in subthreshold oscillations.

Similar to the circuit discussed in section V and the $I_{Na,p}+I_K$ model, the sustained spiking generated by the circuit in Fig. 8 (c) is due to an interplay between the current through the Unipolar memristor (Sodium current) and the MOSFET $Q_1$ (Potassium current). The injected current $I$ charges $C_1$, which causes an increase in $V_{out}$. Therefore, the voltage across the memristor decreases below $V_{rst}$, and the memristor switches to

9TABLE I
COMPARISON OF SPIKING NEURON IMPLEMENTATIONS

| Neuron Circuits | Neuron Model | Number of Components | Signal Gain | Afterhyper-polarization | Subthreshold Oscillations | Biologically Plausible | I/O variables share same node |
|---|---|---|---|---|---|---|---|
| Proposed circuit in Fig. 5 | $I_{Na,p}+I_K$ | 3 FETs, 2 capacitors, 1 resistor | ✓ | ✓ | ✓ | ✓ | ✓ |
| Proposed circuit in Fig. 8 (a) | $I_{Na,p}+I_K$ | 3 FETs, 2 capacitors, 1 resistor | ✓ | ✓ | ✓ | ✓ | ✓ |
| Proposed circuit in Fig. 8 (c) | $I_{Na,p}+I_K$ | 1 FET, 2 capacitors, 1 Unipolar memristor, 1 resistor | ✓ | ✓ | ✓ | ✓ | ✓ |
| [31] | Izhikevich | 14 FETs, 2 capacitors | ✓ | ✓ | ✗ | ✓ | ✓ |
| [28] | HH | 6 FETs, 4 capacitors | ✓ | ✓ | ✗ | ✓ | ✓ |
| [32] | HH | 4 FETs, 2 OTAs, 3 capacitors | ✓ | ✓ | ✗ | ✓ | ✓ |
| [33] | LIF | 34 FETs, 1 capacitor | ✗ | ✗ | ✗ | ✗ | ✗ |
| [34] | LIF | 20 FETs, 1 capacitor | ✓ | ✗ | ✗ | ✗ | ✓ |
| [29] | Morris-Lecar | 8 FETs, 2 capacitors, 1 resistor | ✓ | ✓ | ✗ | ✓ | ✓ |
| [35] | N/A | 12 FETs, 2 capacitors, 1 comparator | ✓ | ✗ | ✗ | ✗ | ✗ |
| [16] | N/A | 2 NbO$_2$ Mott memristors, 3 capacitors, 2 resistors | ✓ | ✓ | ✗ | ✗ | ✗ |
| [17] | N/A | 2 VO$_2$ Mott memristors, 2-3 capacitors, 1-2 resistors | ✓ | ✓ | ✓ | ✗ | ✗ |

the Low Resistance State (LRS). This causes the current through the memristor to increase, which amplifies the increase in $V_{out}$ and results in the upstroke of the spike. During the upstroke, $C_2$ charges and $V_{GS}$ increases. Therefore, the current through $Q_1$'s channel slowly increases, eventually resulting in the spike's downstroke. Consequently, the voltage across the unipolar memristor increases above $V_{rst}$, which switches the memristor back to the HRS state. Due to the slow dynamics of the RC circuit, the membrane potential decreases below the resting potential, resulting in "afterhyperpolarization". After $C_2$ discharges, the injected current starts charging $C_1$ again and the same process is repeated as long as the injected current remains unchanged. This results in the sustained spiking shown in Fig. 8 (d). Therefore, the subthreshold oscillations and spike generation mechanisms of the circuit in Fig. 8 (c) are qualitatively equivalent to that of the circuit in section V and the $I_{Na,p}+I_K$ model discussed in section III. This shows that various NNDR devices/circuits can be used while still obtaining the same qualitative characteristics as that of the $I_{Na,p}+I_K$ model.

VII. DISCUSSION

This work presents a methodology to design minimal neuron circuits that are biologically plausible and use a minimal number of components. That being the case, the $I_{Na,p}+I_K$ model was our model of choice, as it offers both biological plausibility and simplicity. Table 1 compares multiple neuron circuit implementations to the circuits presented in this work. All the circuits proposed use fewer components than the neuron circuit implementations in Table 1. For example, the circuit in Fig. 5 uses only three FETs, two capacitors and one resistor, which are fewer components than any of the circuits in Table 1. Additionally, the circuits presented in this work are all of the resonator-type and are able to exhibit subthreshold oscillations. The ability to exhibit subthreshold oscillations was not demonstrated by any of the implementations in Table 1 except for the implementation from [17]. However, as previously discussed, the neuristor circuit from [17] is not biologically plausible. Furthermore, all the proposed circuits are able to exhibit signal gain and afterhyperpolarization, whilst maintaining biological plausibility. An extensive comparison between our work and other neuron implementations is provided in the Part II companion paper, along with a discussion on prospective applications [19].

VIII. CONCLUSION

In summary, this work aims to present a methodology for designing biologically plausible minimal neuron circuits. This is achieved by introducing several minimal neuron circuits throughout the two parts of the paper. The circuits introduced utilize a minimal number of transistors, memristors and capacitors. In part I, we focus mainly on minimal neurons that act as resonators and highlight the difference between resonators and integrators. Additionally, we show that a more efficient approach to designing neurons is to mimic the $I_{Na,p}+I_K$ model rather than the more complicated HH model. Using this approach, one can achieve neurons showing signal gain, sustained spiking and "afterhyperpolarization" while using a minimal number of components.

Furthermore, we discuss the spike generation mechanism of the $I_{Na,p}+I_K$ model with low threshold activation, and we plot the model's nullclines and bifurcation diagram. We show that the circuits introduced in this work have spike generation mechanisms, nullcline diagrams, and bifurcation diagrams similar to those of the $I_{Na,p}+I_K$ model. Finally, we show that the sodium channel of the $I_{Na,p}+I_K$ model possesses the NNDR property, and we demonstrate how multiple NNDR devices and circuits can be used to achieve the role of a Sodium channel. We provide an example of that by presenting three novel circuits that use NNDR devices/circuits as Sodium channels along with a MOSFET and RC circuit as the Potassium channel. Nevertheless, we have not discussed minimal neurons that act as Integrators. In part II of this work, we thoroughly discuss neurons of the Integrator type, and we propose a method to construct this type of neuron with a minimal number of components [19].